\documentclass{article}

\usepackage{microtype}
\usepackage{amssymb}
\usepackage{graphicx}
\usepackage{subfigure}
\usepackage{booktabs} 

\usepackage{hyperref}

\usepackage[accepted]{icml2019}

\icmltitlerunning{Path-Augmented Graph Transformer Network}

\begin{document}

\twocolumn[
\icmltitle{Path-Augmented Graph Transformer Network}



\begin{icmlauthorlist}
\icmlauthor{Benson Chen}{to}
\icmlauthor{Regina Barzilay}{to}
\icmlauthor{Tommi Jaakkola}{to}
\end{icmlauthorlist}

\icmlaffiliation{to}{Department of EECS, Massachusetts Institute of Technology, Cambridge, USA}

\icmlcorrespondingauthor{Benson Chen}{bensonc@csail.mit.edu}

\icmlkeywords{Machine Learning, Transformer, Chemistry}

\vskip 0.3in
]



\printAffiliationsAndNotice{}  

\begin{abstract}
\setcounter{footnote}{1}

Much of the recent work on learning molecular representations has been based on Graph Convolution Networks (GCN). These models rely on local aggregation operations and can therefore miss higher-order graph properties. To remedy this, we propose Path-Augmented Graph Transformer Networks (PAGTN) that are explicitly built on longer-range dependencies in graph-structured data. Specifically, we use path features in molecular graphs to create global attention layers. We compare our PAGTN model against the GCN model and show that our model consistently outperforms GCNs on molecular property prediction datasets including quantum chemistry (QM7, QM8, QM9), physical chemistry (ESOL, Lipophilictiy) and biochemistry (BACE, BBBP)\footnote{Code to replicate our experiments is provided at \url{https://github.com/benatorc/PA-Graph-Transformer}}.

\end{abstract}
\section{Introduction}

Graph Convolution Networks (GCN) have successfully been applied to molecular graph datasets \citep{duvenaud_conv,kearnes2016molecular,DBLP:journals/corr/NiepertAK16,wengong_chem}. These ``message-passing" algorithms exploit the feature locality of graphs through the usage of convolution operations \citep{message_passing}. However, the convolution operator aggregates only local information, so long-range dependencies are naturally difficult for these models to learn. In molecular graphs, many informative structures are characterized by the paths between nodes. We propose the Path-Augmented Graph Transformer Network (PAGTN) model that utilizes these path features in global attention layers, resulting in a richer, more expressive model. Specifically, our model learns a better representation of the graph in the following ways:

\textbf{Long-range dependencies} In GCNs, long-range dependencies take many convolution layers to learn, because feature aggregation happen only within the immediate neighborhoods of each node. For large enough graphs, GCNs may fail to capture these long-range dependencies entirely. Our PAGTN model can more easily capture these dependencies because every node attends to all other nodes in the graph.

\textbf{Substructures} In graph problems, it is imperative for a model to pick up the important substructures in the graph. GCN models necessitate several layers to propagate information and learn these substructures. The advantage of our model is that this interaction can be learned within a single layer.

We test our PAGTN model against the GCN model on 7 benchmark moelcular property prediction tasks ranging from quantum chemistry (QM7, QM8, QM9), physical chemistry (ESOL, Lipophilictiy) and biochemistry (BACE, BBBP) \citep{wu2018moleculenet}. Each dataset focuses on a different property of the molecule, making composition of these datasets highly variable. Nevertheless, our model consistently shows improved performance against the GCN baseline, demonstrating that our model can learn more powerful representations.
\section{Related Works}

Transformer architectures have triumphed over traditional recurrent and convolution models in many natural language tasks such as machine translation \citep{vaswani_transformer}. While recurrent and convolution models often incorporate a single attention layer at the top \citep{luong2015effective}, it has been shown that using only these globally-connected self-attention layers learns a much more powerful model. 

Attention models on graphs have been explored in previous works. Primarily, the Graph Attention Network \citep{graph_attn} and its variants \cite{gong2018adaptive, DBLP:journals/corr/abs-1803-07294,DBLP:journals/corr/abs-1806-00770} aggregates information within local neighborhoods by using attention. We emphasize that our model focuses on the global connectivity of the nodes. Moreover, our model does not use any complex attention mechanism across layers, but rather provides a simple framework using the path features that works well empirically. Another proposed model, Graph Transformer \citep{li2019graph}, uses global attention layers, but that model does not extend to graphs in which edge and path features are important.
\section{Model}

In this section, we first briefly overview the Transformer model. Then, we will go over our contributions, describing our variant of the Transformer model that uses path features to learn expressive representations of graphs.

\subsection{Transformer}
The Transformer model \citep{vaswani_transformer}, in contrast to traditional recurrent or convolution architectures, consists of fully-connected attention layers. These models use multi-head self-attention, which confers more flexibility for the attention module. The attention layers are connected by position-wise feed-forward layers, with residual links and layer normalization present at each layer. 

The transformer model itself has no direct notion of relative position, so it uses positional encodings in the form of sinusoidal functions. However, this form of positional encoding is not possible in graphs, because there is no longer a natural sequential ordering of the nodes. We introduce path features, which represent how two nodes are connected. These path features influence the attention module in the network, so that the node embeddings are globally aware. We first explain how we construct these path features, then how they are incorporated into the attention framework.

\subsection{Path Features}

We compute the path features between each node pair by taking the shortest path between them. Due to cycles on graphs, these shortest paths may not be unique. For molecular graphs, these cycles arise due to ring substructures on the graphs. Because the edge features are consistent within a single ring or cycle, multiple paths are almost always equivalent feature-wise; therefore, this approach is sensible for our model.

For efficiency, we truncate the path features between nodes up to a distance $d$ apart. We make the assumption that as the distance between two nodes increases, the connectivity between the two nodes matter less. Therefore, this constraint puts a natural regularizer on the model. So while each node attend to all other nodes in the graph, that node only has rich edge features for a local neighborhood.

The path features between two nodes $i \rightarrow j$ is a concantenation of the following three components:

\textbf{Edge features}: are constructed by concatenating the individual bond features of the shortest path between $i \rightarrow j$. Let $b_k$ be the bond features of the $k$th bond along the path, which includes the bond type, conjugacy and ring membership (whether or not that bond is in a ring) features. Then, the edge features are just the concatenation of the features: $[b_1; b_2; ...; b_n]$. Note that if $n > d$, we zero out these features, and if $n < d$, we pad the feature vector with zeros.

\textbf{Distance}: is a one-hot feature of the distance between two nodes $i \rightarrow j$, truncated by  $d$.

\textbf{Ring Membership}: is a one-hot feature denoting whether the node $i$ and node $j$ are in the same ring. For molecular graphs, we find that it's also helpful to include one-hot features for specific rings such as five/six-membered aromatic rings. Note that this is distinct from the bond ring membership features which indicates whether a particular bond is part of a ring.

\begin{figure}[ht]
\vskip 0.2in
\begin{center}
\centerline{\includegraphics[width=\columnwidth]{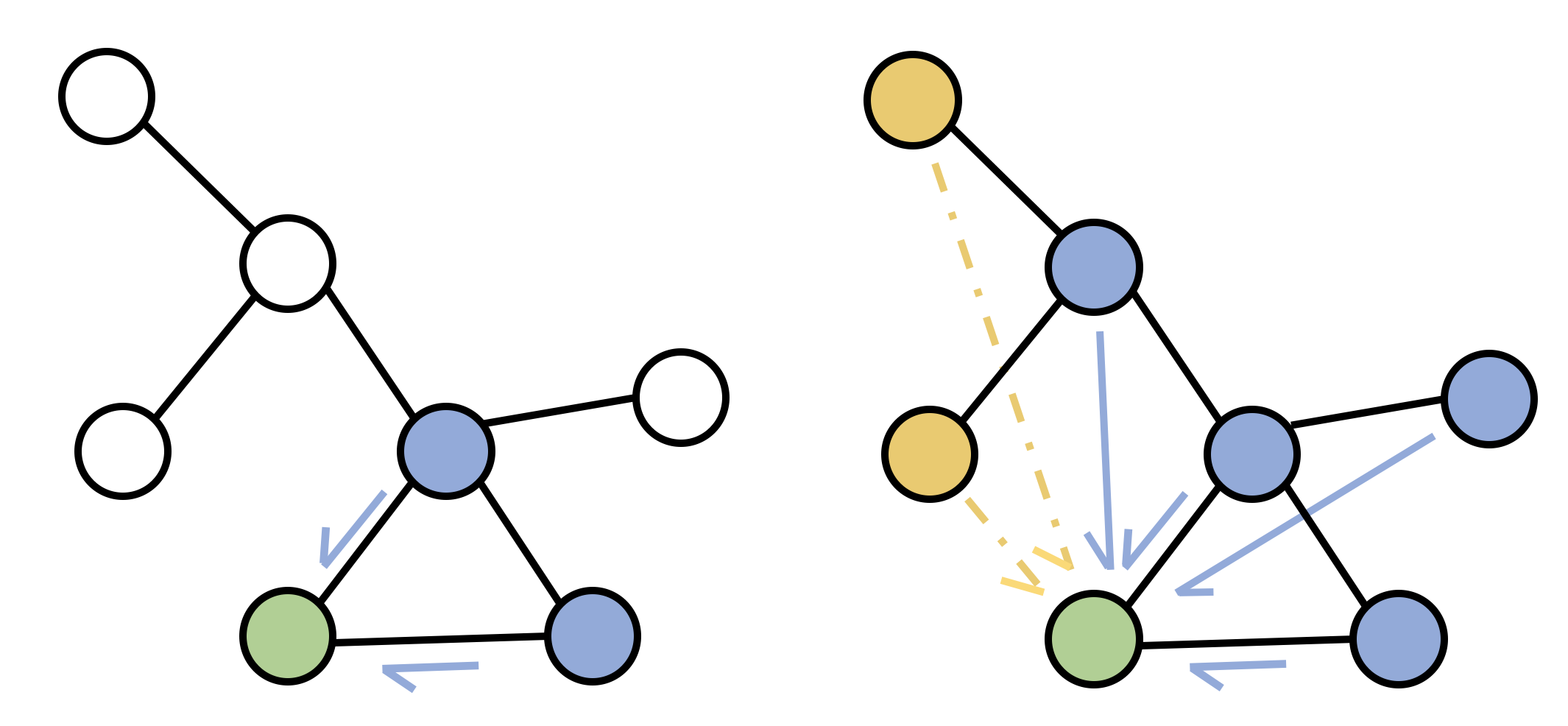}}
\caption{Illustration of graph propagation properties for GCN (left) and our PAGTN model (right). For the GCN, the source attention node (green) only attends to its immediate neighbors (blue). In the PAGTN, the source attention node (green) has connectivity information in the form of path features for its local neighborhood, $d=2$, (blue), but also attends to all other nodes (yellow).}
\label{graph_prop}
\end{center}
\vskip -0.2in
\end{figure}






A comparison of the information propagation properties of the network layers is illustrated in Figure \ref{graph_prop}. In regular GCNs, only the direct neighborhood is impacted--which can require many layers of computation to learn from the graph. In our PAGTN model, every node is globally connected, which makes learning complex dependencies easier.

\subsection{Additive Self-Attention}
Although transformer models normally use scaled dot-product attention, we found in our experiments that an additive form of attention was easier to train and resulted in better performance. One way we deviate from standard self-attention modules is that we exclude the source node when computing attention for that node. The residual links at each layer grounds the learned embedding at each layer to be representative of the original input node. 

Define $\mathbf{x} = (x_1, x_2, ... ,x_n) \in \mathbb{R}^{n \times F_n}$ as a matrix of the input node features, where $n$ is the number of nodes and $F_n$ is the number of node features. Similarly, let $\mathbf{p} = (p_{1, 1}, p_{1, 2}... p_{n, n}) \in \mathbb{R}^{n \times n \times F_p}$ be a matrix of the input pairwise path features where $F_p$ is the number of path features.

At each layer, we update the node features by computing a weighted average using learned attention weights. Let $\mathbf{h}^l = (h^l_1, h^l_2, ... ,h^l_n) \in \mathbb{R}^{n \times F_m}$ represent the node features at layer $l$, where $F_m$ is the number of model features. Note that the elements of $\mathbf{h}^0$ are the linearly transformed input features ($\mathbf{h^0} = W \mathbf{x^T}$). We compute $s^l_{i, j}$, the attention score of node $i \rightarrow j$, as:

\begin{equation}
    s^{l}_{i, j} = W^{S_2} \Big[\textrm{LeakyReLU} \Big(W^{S_1} [h^{l-1}_i; h^{l-1}_j; p_{i, j}]\Big)\Big]
\end{equation}

The attention probabilities ${a_{i, j}}$ are calculated as a softmax over the attention scores. As mentioned earlier, we exclude the source node itself when computing the attention probabilities.

\begin{equation}
    \alpha^{l}_{i, j} = \textrm{softmax} (s^l_{i, j}) = \frac{\exp(s^l_{i, j})}{\sum_{j' \neq i} \exp(s^l_{i, j'})}
\end{equation}

Using attention probabilities, we can compute a weighted average over the node features. Since we note the importance of path features in graphs, we define the output features to be a function of both node and path features. Here, $\sigma$ is some non-linear function (we use ReLU for our experiments).

\begin{equation}
    h^{l}_i = \sigma \big( W^{H_2} h^{l-1}_i + \sum_{j \neq i} \alpha^l_{i, j} W^{H_1} [h^{l-1}_j; p_{i, j}]\big)
\end{equation}

As introduced in \citep{vaswani_transformer}, multi-head attention can often benefit the model by allowing it more easily to attend to different aspects of the input data. If we split the attention into $K$ heads, we can define the update rule for $h^{l}_i$ as a function of the embeddings associated with invididual heads $h^{l, k}_i$:

\begin{equation}
    h^{l}_i = \Big\|_k \sigma \big( W^{H_2, k} h^{l-1, k}_i + \sum_{j \neq i} \alpha^{l, k}_{i, j} W^{H_1, k} [h^{l-1, k}_j; p_{i, j}]\big)
\end{equation}

Here, $\|$ is the concatenation operator. Empirically, we find that using multi-head attention helps on some tasks, but not on all tasks.

\subsection{Molecule Embedding}
Since we are interested in property prediction tasks for the molecule as a whole, we compute a molecule embedding $h_M$ by aggregating the individual node embeddings. Here, we add a residual link to the input features, $\textbf{x}$, of the network.

\begin{equation}
    h_M = \sum_i \sigma \Big(W^M [h^L_i; x_i] \Big)
\end{equation}

We choose the sum operator to aggregate the feature embeddings, which has higher expressive power than other classic operators \citep{graph_power}. The target property is predicted using a 1-layer MLP with $h_M$ as input.
\section{Experiments}

\subsection{Experimental Setup}
We test our model on 7 benchmark property prediction tasks, including quantum mechanics (QM7, QM8, QM9), physical chemistry (ESOL, Lipophilicity) and biochemistry (BACE, BBBP) \citep{wu2018moleculenet}.

We split each dataset into 10 different folds of 80:10:10 (train:validation:test) splits, and record the average performance over the folds using the appropriate measure for each dataset. Since these datasets feature markedly different properties, we tune the hyperparameters of the model for individual datasets.

\subsection{Baselines}
We compare our transformer to several baselines.

\textbf{MolNet} Molecule Net \citep{wu2018moleculenet} tested many graph-based deep learning methods as well as more conventional methods on these property prediction datasets. We use their top performing model for each dataset.

\textbf{GCN} This is a traditional graph convolution model, and here we use a similar model to \citep{wengong_chem}. We find that this model achieves very competitive results compared MolNet (which itself uses many different graph-based convolution models), and therefore is a fair baseline. GCN models can have a self-attention layer at the top, but we find empirically that this often hurts performance so we do not include this attention layer in our baseline.

\textbf{PAGTN (Local)} We include a variant of our PAGTN model, which does not attend to nodes for which there are no path features. That is, the model masks out nodes that are further than $d$ from the source attention node. We include this baseline to show that global attention does indeed improve performance.

Our proposed model is dubbed the \textbf{PAGTN (Global)}, which attends globally to all nodes.

\begin{table*}[t]
\setlength\tabcolsep{2.5 pt}
\caption{Results comparing our PAGTN model to various baselines. The metrics used were MAE for the quantum mechanics datasets (QM7, QM8, QM9), RMSE for the physical chemistry datasets (ESOL, Lipophilicity), and AUC for the biochemistry datasets (BACE, BBBP). The bold numbers represent the model with the best performance.}
\label{prop_table}
\vskip 0.15in
\begin{center}
\begin{small}
\begin{sc}
\begin{tabular}{lrrlrlrlrl}
\toprule
Data set & \# Data  & Metric & MolNet & \multicolumn{2}{c}{GCN} & \multicolumn{2}{c}{PAGTN (Local)} & \multicolumn{2}{c}{PAGTN (Global)}  \\
\midrule
QM7 & 6,830 & MAE & --\footnotemark & 52.4 & $\pm$ 2.8& 48.9 & $\pm$ 3.4& \textbf{47.8} & \textbf{$\pm$ 3.0} \\
QM8 & 21,786 & MAE & .0143 & .0105 & $\pm$ .0003 & .0108 & $\pm$ .0003 & \textbf{.0102} & $\pm$ \textbf{.0003}\\
QM9 & 133,885 & MAE & 2.35 & 2.20 & $\pm$ .03 & 2.10 & $\pm$ .04& \textbf{2.07} & $\pm$ \textbf{.05} \\
\midrule
ESOL & 1,128 & RMSE & .580 & .587 & $\pm$ .05 & .592 & $\pm$ .06 & \textbf{.554} & $\pm$ \textbf{.06} \\
Lipophilicity & 4,200 & RMSE & .655 & .578 & $\pm$ .05 & .592 & $\pm$ .05 & \textbf{.572} & $\pm$ \textbf{.04} \\
\midrule
BACE & 1,513 & AUC & .867 & .878 &$\pm$ .02& .876 &$\pm$ .02 & \textbf{.880} & \textbf{$\pm$ .01}\\
BBBP & 2,039 & AUC & .729 & .907 & $\pm$ .03 & .898 & $\pm$ .04 & \textbf{.913} & \textbf{$\pm$ .03} \\
\bottomrule
\end{tabular}
\end{sc}
\end{small}
\end{center}
\vskip -0.1in
\end{table*}

\subsection{Property Prediction}

The results of the property prediction tasks can be seen from Table \ref{prop_table}. We first see that the GCN model is very comparable to those of MolNet \citep{wu2018moleculenet}. And compared to the GCN model, our PAGTN model achieves surperior performance in all 7 of these property prediction tasks, illustrating the broad representational power of the model. Furthermore, we see from the local PAGTN model that by attending globally rather than restricting to the local neighborhood, we always see an improvement in performance. This reveals that the global attention does indeed help the model.

\footnotetext{MolNet uses a stratified sampling of the data for QM7, whereas we use random sampling for this work.}

\subsection{Ring Membership}

To help elucidate why the PAGTN formulation is better than that of GCN, we turn to a synthetic task. We note that certain properties such as ring membership can prove difficult for regular graph convolution networks. To test this observation, and to demonstrate the effectiveness of our PAGTN model, we create a synthetic dataset by choosing a subset of 5,769 molecules from the property prediction datasets that have at least 2 rings. For each molecule, we randomly choose 5 pairs of atoms that are in the same ring, and 5 pairs of atoms that are in different rings. For atoms in fused ring systems, we count two atoms in the same ring if they are in the smallest possible ring system.

\begin{table}[t]
\caption{Results comparing the GCN and the PAGTN (Global) models on a synthetic ring membership prediction task, which is to test whether or not two nodes are in the same ring on the graph. GCN does cannot always predict this property well, while the PAGTN can easily incorporate these features into the model.}
\label{ring_table}
\vskip 0.15in
\begin{center}
\begin{small}
\begin{sc}
\begin{tabular}{lcc}
\toprule
Model & Accuracy & AUC   \\
\midrule
GCN & 91.6 & 96.5 \\
PAGTN (Global) & 97.8 & 99.8 \\
\bottomrule
\end{tabular}
\end{sc}
\end{small}
\end{center}
\vskip -0.1in
\end{table}

\begin{figure}[ht]
\vskip 0.2in
\begin{center}
\centerline{\includegraphics[width=0.7\columnwidth]{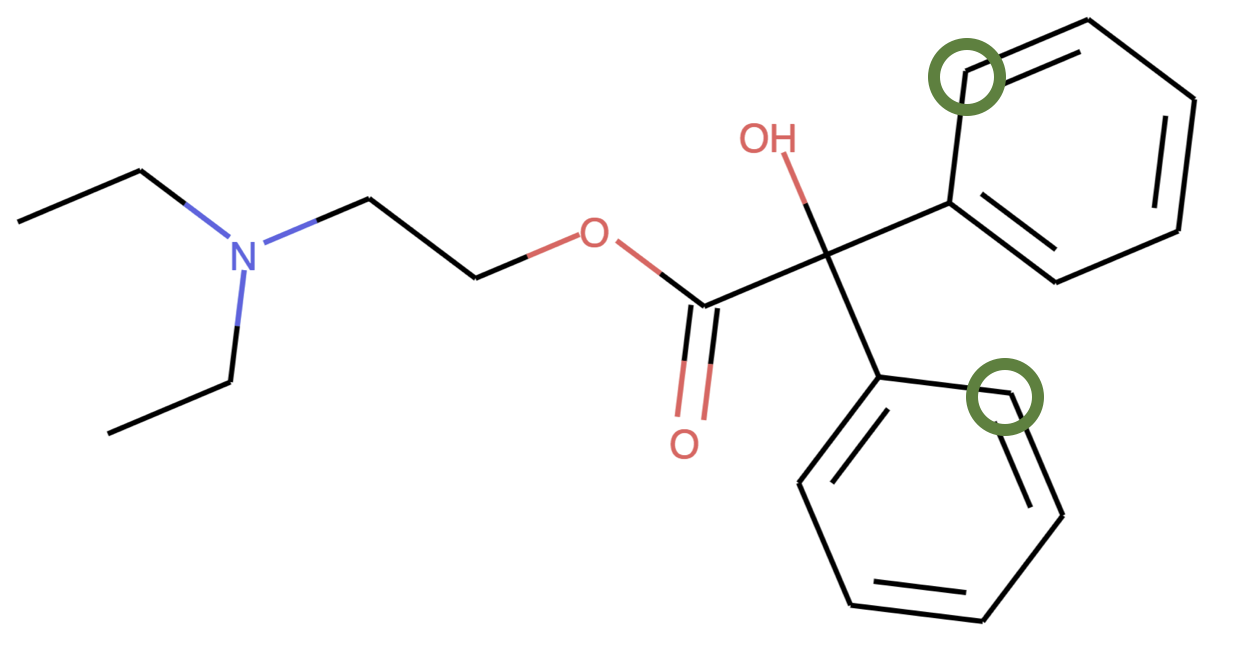}}
\caption{The two green-circled atoms are completely symmetric, so their output feature embeddings are equivalent. Since the ring membership prediction is made by aggregating pairwise node features, it is impossible to tell whether any other atom is in the same or different ring from these two atoms.}
\label{sym-fig}
\end{center}
\vskip -0.2in
\end{figure}

From Table \ref{ring_table}, we see that the GCN fails to perfectly predict ring membership. This is not surprising as the convolution operation has to learn to disambiguate features of nodes in same and different rings. These subtle but important graph features are imperative for models to fully capture the representation of the graph. Our PAGTN naturally solves this issue, since we can incorporate these features as a part of the network, whereas it is a lot more difficult to incorporate these features in the local convolution model. Note that the PAGTN still does not solve the problem perfectly, and this is due to the fact that in highly symmetrical graphs, multiple nodes are equivalent which leads to ambiguous ring membership as see from Figure \ref{sym-fig}.

\section{Conclusion}

In this paper, we introduced the PAGTN model that exploits the connectivity structure of the data in its global attention mechanisms. Through the path features that we engineer into model's attention layers, our model better captures the complex structures of graphs compared to GCNs. On 7 different chemical property prediction tasks, we have shown that our PAGTN model can outperform traditional GCNs, and we hope that these global-attention models that incorporate path features will be used more frequently in works on molecular graphs moving forward. 

\bibliography{ref}
\bibliographystyle{icml2019}

\end{document}